\documentclass[final,leqno]{siamltex}

\usepackage{amsmath,amssymb,amsfonts}
\usepackage{algorithm}
\usepackage{algorithmic}
\usepackage{multirow}
\usepackage{booktabs}
\usepackage{graphicx}

\DeclareMathOperator{\prox}{prox}

\title{Practical Algorithms for Learning Near-Isometric Linear Embeddings\thanks{This research was conducted as part of the California Research Training Program in Computational and Applied Mathematics 2014.}} 

\author{Jerry Luo\footnotemark[2] \and Kayla Shapiro\footnotemark[3] \and Hao-Jun Michael Shi\footnotemark[4] \and Qi Yang\footnotemark[5] \and Kan Zhu\footnotemark[6]}

\begin{document}
\maketitle

\renewcommand{\thefootnote}{\fnsymbol{footnote}}

\footnotetext[2]{Department of Mathematics, University of Arizona, Tucson, AZ 85721. (jerryluo@math.arizona.edu)}
\footnotetext[3]{Department of Computing, Imperial College London, London SW7 2AZ, United Kingdom. (k.shapiro@berkeley.edu)}
\footnotetext[4]{Department of Mathematics, University of California, Los Angeles, Los Angeles, CA 90024. (hjmshi@ucla.edu)}
\footnotetext[5]{Department of Mathematics, University of Southern California, Los Angeles, 90089. (yangq@usc.edu) }
\footnotetext[6]{Department of Computer Science, Columbia University, New York, NY 10027. (kzhu9@ucla.edu)}

\slugger{siopt}{xxxx}{xx}{x}{x--x}

\begin{abstract} 
We propose two practical non-convex approaches for learning near-isometric, linear embeddings of finite sets of data points. Given a set of training points $\mathcal{X}$, we consider the secant set $S(\mathcal{X})$ that consists of all pairwise difference vectors of $\mathcal{X}$, normalized to lie on the unit sphere. The problem can be formulated as finding a symmetric and positive semi-definite matrix $\boldsymbol{\Psi}$ that preserves the norms of all the vectors in $S(\mathcal{X})$ up to a distortion parameter $\delta$. Motivated by non-negative matrix factorization, we reformulate our problem into a Frobenius norm minimization problem, which is solved by the Alternating Direction Method of Multipliers (ADMM) and develop an algorithm, \textit{FroMax}. Another method solves for a projection matrix $\boldsymbol{\Psi}$ by minimizing the restricted isometry property (RIP) directly over the set of symmetric, postive semi-definite matrices. Applying ADMM and a Moreau decomposition on a proximal mapping, we develop another algorithm, \textit{NILE-Pro}, for dimensionality reduction. FroMax is shown to converge faster for smaller $\delta$ while NILE-Pro converges faster for larger $\delta$. Both non-convex approaches are then empirically demonstrated to be more computationally efficient than prior convex approaches for a number of applications in machine learning and signal processing. 
\end{abstract} 


\pagestyle{myheadings}
\thispagestyle{plain}

\section{Introduction}

\subsection{Motivation}

The size of the acquired raw data currently poses a challenge to current state-of-the-art information processing systems. Since many machine learning algorithms' computational efficiency scale with the complexity of the data, machine learning researchers have introduced a family of algorithms for \textit{dimensionality reduction} to address this issue. Dimensionality reduction algorithms construct a concise representation of high-dimensional data on a lower-dimensional subspace, with as minimal loss of intrinsic information as possible. This representation is often referred to as a low-dimensional \textit{embedding}.

The canonical approach in statistics for constructing a linear embedding is principal components analysis (PCA) \cite{moore1981principal}. PCA is a linear embedding technique that projects data points onto a lower-dimensional subspace spanned by the principal components that contain the most variability within the data. PCA benefits from being computationally efficient and easily generalizable to new data sets; however, it fails to preserve pairwise distances between sample data points, sometimes rendering two distinct points indistinguishable in the low-dimensional embedding space. This can potentially hamper the performance of PCA and other similar algorithms.

Other popular nonlinear, manifold learning methods, such as ISOMAP and locally linear embedding (LLE), preserve geometric structure by approximating geodesics from $k$-nearest neighbors. Although these methods produce embeddings which are easy to explicitly store and generalize, most fail to preserve all pairwise distances between data points. 

A linear embedding technique that preserves all pairwise distances is the method of \textit{random projections}. Given $\mathcal{X}$, a cloud of $Q$ data points in a high-dimensional Euclidean space $\mathbb{R}^N$, the Johnson-Lindenstrauss Lemma \cite{johnsonlindenstrauss} states that there exists a linear, near-isometric, or distance preserving, embedding such that $\mathcal{X}$ can be mapped to a subspace of dimension $M = \mathcal{O}(\log Q)$ with high probability. Despite its conceptual simplicity, random projections suffers from probabilistic and asymptotic theoretical guarantees in Johnson-Lindenstrauss \cite{johnsonlindenstrauss}. A random projections mapping is also independent of the data under consideration, failing to utilize the geometric structure of the data. 

We would like to deterministically devise a linear embedding for dimensionality reduction that preserves all normalized pairwise distances. In particular, linear embeddings can be explicitly stored using a matrix operator and can therefore be quickly applied to any new data point.

\subsection{Related Work}

Using the geometric structure of the data, Hegde, et. al. developed a new deterministic approach, NuMax, to construct a near-isometric, linear embedding \cite{HegdeSankaranarayananYinBaraniuk2015_numax}. Given a training set $\mathcal{X} \subset \mathbb{R}^N$, the \textit{secant set} is constructed by taking all pairwise difference vectors of $\mathcal{X}$, which are then normalized to lie on the unit sphere. Hegde, et. al. formulated a rank minimization problem with affine constraints to construct a \textit{projection matrix} $\psi$ that preserves norms of all vectors in $S(\mathcal{X})$ up to a distortion parameter $\delta$. They then relax this problem to a convex program that can be solved using a tractable semidefinite program (SDP), with column generation for large data sets, and develop NuMax based on the Alternating Direction Method of Multipliers (ADMM). This framework deterministically produces a near-isometric linear embedding. Other algorithmic approaches for finding near-isometric linear embeddings are also described in \cite{grant2013nearly,hegde2012near,sadeghian2013energy}.

\subsection{Organization}

The rest of the paper is organized as follows. We review the restricted isometry property and NuMax algorithm in \S \ref{background}. The FroMax algorithm is introduced in \S \ref{fromax}. NILE-Pro is discussed in \S \ref{nile-pro}. Rank adjustment and column generation methods which increase computational efficiency for large data sets is introduced in \S \ref{racg}. Numerical simulations and runtime performance results are presented in \S \ref{experiments}. Lastly, \S \ref{discussion} concludes the paper and gives direction for future work.

\section{Background} \label{background}

\subsection{Restricted Isometry Property (RIP)}

E. Candes, et. al. introduced a formal, relaxed notion of isometry in \cite{candes2005decoding} as follows:\\

\begin{definition} 
Suppose $M \leq N$ and consider $\mathcal{X} \subset \mathbb{R}^N$. An embedding operator $\mathcal{P} : \mathcal{X} \rightarrow \mathcal{R}^M$ satisfies the \textit{restricted isometry property (RIP)} on $\mathcal{X}$ if there exists a positive constant $\delta > 0$ such that, for every $\mathbf{x}, \mathbf{x'} \in \mathcal{X}$, the following relation holds:
\begin{equation}
(1 - \delta)\| \mathbf{x} - \mathbf{x'} \|_2^2 \leq \| \mathcal{P}\mathbf{x} - \mathcal{P}\mathbf{x'} \|_2^2 \leq (1 + \delta)\| \mathbf{x} - \mathbf{x'} \|_2^2.
\end{equation}
\end{definition}

We may also refer to $\delta$ as the \textit{isometry constant}. Intuitively, this notion of near-isometry requires the distance of every pair of points in $\mathcal{X}$ to be almost preserved. Hegde, et. al. \cite{HegdeSankaranarayananYinBaraniuk2015_numax} leveraged this idea to develop a framework for finding low rank matrices that satisfy the RIP.

\subsection{NuMax}
In this section, we review Hegde et. al. \cite{HegdeSankaranarayananYinBaraniuk2015_numax}'s work on NuMax. Given a data set $\mathcal{X} \subset \mathbb{R}^N$, Hegde et. al. formulate the secant set as follows:
\begin{equation}
S(\mathcal{X}) = \left\{ \dfrac{\mathbf{x} - \mathbf{x'}}{\|\mathbf{x} - \mathbf{x'}\|_2} , \mathbf{x, x'} \in \mathcal{X}, \mathbf{x} \neq \mathbf{x'} \right \}.
\end{equation}
 Hegde, et. al. seeks to find a projection matrix $\boldsymbol{\Psi} \in \mathbb{R}^{M \times N}$ with the smallest possible rank that satisfies the RIP on $S(\mathcal{X})$ for a given $\delta > 0$. This problem is then cast as an optimization problem over all symmetric matrices which we denote as $\mathbb{S}^{N \times N}$. Let $\mathbf{P} = \boldsymbol{\Psi}^T\boldsymbol{\Psi} \in \mathbb{S}^{N \times N}$ with $\operatorname{rank}(\mathbf{P}) = M$. Then for all secants $\mathbf{v}_i \in S(\mathcal{X})$, we may rewrite the RIP constraint as:
 \begin{equation}
(1 - \delta)\| \mathbf{v}_i \|_2^2 \leq \| \boldsymbol{\Psi} \mathbf{v}_i \|_2^2 \leq (1 + \delta)\|\mathbf{v}_i\|_2^2.\\
\end{equation}
After grouping we get: 
\begin{equation}
| \| \boldsymbol{\Psi} \mathbf{v}_i \|_2^2 - \| \mathbf{v}_i\|_2^2 | \leq \delta\|\mathbf{v}_i\|_2^2.\\
\end{equation}
Then we isolate $\delta$ and use the fact that $ \| \boldsymbol{\Psi} \mathbf{v}_i \|_2^2 = \mathbf{v}_i^T \mathbf{P} \mathbf{v}_i$:
\begin{align}
| \| \boldsymbol{\Psi} \mathbf{v}_i \|_2^2 - 1 | & \leq \delta\\
|\mathbf{v}_i^T \mathbf{P} \mathbf{v}_i - 1 | & \leq \delta.
\end{align}
Let $\boldsymbol{1}_S$ denote the $S$-dimensional ones vector and $\mathcal{A}: \mathbb{R}^{N \times N} \to \mathbb{R}^S$ where $S=|S(\mathcal{X})|$, and $\mathcal{A}:X \mapsto \{ \mathbf{v}_i^T X \mathbf{v}_i\}_{i = 1}^S$, where the output is understood to be an $S$ dimensional vector with the $i$-th entry being $\mathbf{v}_i^T X \mathbf{v}_i$. This admits the rank minimization problem:
\begin{equation}
\begin{aligned}
& \underset{P}{\text{minimize}}
& & \operatorname{rank}(\mathbf{P}) \\
& \text{subject to}
& & \|\mathcal{A}(\mathbf{P}) - \boldsymbol{1}_S\|_\infty \leq \delta\\ 
& & & \mathbf{P} \succeq 0.
\end{aligned}
\end{equation} 
Here, $\mathbf{P} \succeq 0$ means $\mathbf{P}$ is symmetric positive semidefinite. However, since rank minimization is a non-convex, NP-hard problem \cite{recht2010guaranteed}, a convex relaxation is performed on the objective to obtain the following nuclear-norm minimization:
\begin{equation}
\begin{aligned}
&\underset{\mathbf{P}}{\text{minimize}}
& & \|\mathbf{P}\|_* \\
& \text{subject to}
& & \|\mathcal{A}(\mathbf{P}) - \boldsymbol{1}_S\|_\infty \leq \delta \\
& & & \mathbf{P} \succeq 0,
\end{aligned}
\end{equation}
where $\| \mathbf{P} \|_*$ is the nuclear norm of $\mathbf{P}$, or the sum of the singular values of $\mathbf{P}$. Then the desired linear embedding $\boldsymbol{\Psi} \in \mathbb{R}^{M \times N}$ can be found by taking a matrix square root of the minimizer $\mathbf{P}^* = \mathbf{U} \boldsymbol{\Gamma} \mathbf{U}^T$ by
\begin{equation}
\boldsymbol{\Psi} = \boldsymbol{\Gamma}_M^{1/2} \mathbf{U}_M^T,
\end{equation}
where $\boldsymbol{\Gamma}_M = \operatorname{diag}\{\boldsymbol{\lambda}_1, ..., \boldsymbol{\lambda}_M\}$ denotes the M leading (non-zero) eigenvalues of $\mathbf{P}^*$ and $\mathbf{U}_M$ are the corresponding eigenvectors. 

Applying the Alternating Direction Method of Multipliers (ADMM), the optimization problem is rewritten by introducing auxiliary variables $\mathbf{L} \in \mathbb{S}^{N \times N}$ and $\mathbf{q} \in \mathbb{R}^S$:
\begin{equation}
\begin{aligned}
&\underset{\mathbf{P, L, q}}{\text{minimize}}
& & \|\mathbf{P}\|_* \\
& \text{subject to}
& & \mathbf{P} = \mathbf{L}\\
& & & \mathcal{A}(\mathbf{L}) = \mathbf{q}, \\
& & &\|\mathbf{q} - \boldsymbol{1}_S\|_\infty \leq \delta\\ 
& & & \mathbf{P} \succeq 0.
\end{aligned}
\end{equation}
Introducing auxiliary variables $\mathbf{L}$ and $\mathbf{q}$ allows constrained optimization techniques to be applied while breaking up the problem into simpler subproblems. The linear constraints are then relaxed to form an augmented Lagrangian as follows:
\begin{equation}
L_A(\mathbf{P, L, q}; \boldsymbol{\Gamma}, \boldsymbol{\omega}) = \|\mathbf{P}\|_* + \frac{\beta_1}{2} \|\mathbf{P} - \mathbf{L} - \boldsymbol{\Gamma}\|_F^2 + \frac{\beta_2}{2}\|\mathcal{A}(\mathbf{L}) - \mathbf{q} - \boldsymbol{\omega}\|_2^2,
\end{equation}
where $\boldsymbol{\Gamma} \in \mathbb{S}^{N \times N}$ and $\boldsymbol{\omega} \in \mathbb{R}^S$ represent the scaled Lagrange multipliers and $\beta_1, \beta_2 \in \mathbb{R}$ are chosen parameters.

NuMax then solves the following augmented Lagrangian problem:
\begin{equation}
\begin{aligned}
&\underset{\mathbf{P}, \mathbf{L}, \mathbf{q}, \boldsymbol{\Gamma}, \boldsymbol{\omega}}{\text{minimize}}
& & L_A(\mathbf{P, L, q}, \boldsymbol{\Gamma}, \boldsymbol{\omega})\\
& \text{subject to}
& & \|\mathbf{q} - \boldsymbol{1}_S\|_\infty \leq \delta\\
& & & \mathbf{P} \succeq 0,
\end{aligned}
\end{equation}
$\mathbf{P,L}$ and $\mathbf{q}$ are optimized in an alternating fashion, i.e. optimized one at a time with the others held fixed. This optimization can then be solved by three easier sub-problems, admitting a computationally efficient solution. 

For more information regarding theoretical and empirical properties of NuMax, please refer to Hegde et. al. \cite{HegdeSankaranarayananYinBaraniuk2015_numax}.

This framework, though slower than conventional methods such as PCA and random projections, deterministically admits a projection matrix satisfying the RIP with low rank. However, NuMax computes a singular value decomposition of $\mathbf{P}$ each iteration, which is computationally expensive. Furthermore, though minimizing the nuclear-norm tends to give low rank matrices, NuMax does not theoretically guarantee the lowest rank embedding for a given $\delta$. 

These issues motivate the pursuit of other practical algorithms that optimize similar non-convex problems that may admit low rank, near-isometric projection matrices that give faster, but sufficient (not necessarily optimal) results. Rather than solving both the rank minimization and near-isometry problems simultaneously, we solve a simpler non-convex problem quickly to find a near-isometric projection matrix and apply a rank adjustment heuristic to choose a minimal rank.

\section{FroMax} \label{fromax}

Our first algorithm, \textit{Frobenius norm minimization with Max-norm constraints}, or \textit{FroMax}, combines ideas from NuMax and NMF to formulate a Frobenius norm minimization problem which we then solve based on ADMM, similar to NuMax \cite{XuYinWenZhang2012_alternating}. Though this algorithm does not discover the optimal rank for $\boldsymbol{\Psi}$, we combine FroMax with a rank adjustment heuristic to find low rank embeddings.

\subsection{Non-Negative Matrix Factorization (NMF)}

FroMax is motivated by ideas from \textit{non-negative matrix factorization}. Non-negative matrix factorization (NMF) is a group of algorithms that factorize a non-negative matrix $\mathbf{V}$ into two low-rank non-negative matrices $\mathbf{W}$ and $\mathbf{H}$ \cite{NIPS2000_1861}. More rigorously, let $\mathbf{V} \in \mathbb{R}^{N \times M}$ be given, then we solve for $\mathbf{W} \in \mathbb{R}^{M \times Q}$, and $\mathbf{H} \in \mathbb{R}^{Q \times N}$ by solving the following optimization problem:
\begin{equation}
\begin{aligned}
&\underset{\mathbf{W, H}}{\text{minimize}}
& & \|\mathbf{WH - V} \|_F^2\\
& \text{subject to}
& & W_{ij} \geq 0, H_{ij} \geq 0, \forall i,j. \\
\end{aligned}
\end{equation}
NMF motivates the problem formulation for our first algorithm, FroMax.

\subsection{Optimization Framework}

We formulate a specialized matrix factorization minimization problem to solve for a near-isometric linear embedding as follows: 
Given a desired rank $r << N$, let $\boldsymbol{\Psi} \in \mathbb{R}^{r \times N}$ be the variable projection matrix. Here, we seek to solve:
\begin{equation}\label{eq:fromax}
\begin{aligned}
& \underset{\mathbf{P}, \boldsymbol{\Psi}}{\text{minimize}}
& & \frac{1}{2}\|\mathbf{P} - \boldsymbol{\Psi}^T \boldsymbol{\Psi} \|_F^2 \\
& \text{subject to}
& & \| \mathcal{A}(\mathbf{P}) - \boldsymbol{1}_S\|_\infty \leq \delta,
\end{aligned}
\end{equation}
where $\mathbf{P} \in \mathbb{R}^{N \times N}$ is a variable.

We introduce auxiliary variables to simplify our optimization and break up our problem into simpler subproblems that can be solved by applying ADMM. In particular, let $\mathbf{Y} = \boldsymbol{\Psi} \in \mathbb{R}^{r \times N}$, $\mathbf{X} = \mathbf{Y}^T \in \mathbb{R}^{N \times r}$. Then
\begin{equation}\label{eq:fromax_admm}
\begin{aligned}
& \underset{\mathbf{P,X,Y,q}}{\text{minimize}}
& & \frac{1}{2}\|\mathbf{P - XY}\|_F^2 \\
& \text{subject to}
& &  \mathcal{A}(\mathbf{P}) = \mathbf{q}\\ 
& & & \mathbf{Y} = \mathbf{X}^T\\
& & & \| \mathbf{q} - \boldsymbol{1}_S\|_\infty \leq \delta.
\end{aligned}
\end{equation}
This gives $\mathbf{Y} = \boldsymbol{\Psi} \in \mathbb{R}^{r \times N}$ such that the RIP holds for all secant vectors in the secant set $S(\mathcal{X})$ for an isometry constant $\delta$. The optimization formulation for (\ref{eq:fromax}) is conceptually simple, only requiring the input data set $\mathcal{X}$, desired isometry constant $\delta > 0$ and desired rank $r$.

An important caveat is that our optimization problem is non-convex. Thus, we cannot guarantee that FroMax will converge to the optimal solution of (\ref{eq:fromax}). However, various experiments in \S \ref{experiments} indicate that FroMax yields excellent, stable results for real-world data sets and finds projection matrices faster than NuMax and other convex approaches. We implement ADMM since Wang et. al. \cite{wang2015global} indicate that ADMM is more likely to converge than the Augmented Lagrangian Method for nonconvex, nonsmooth problems.

\subsection{ADMM}

We develop our algorithm, FroMax, to solve (\ref{eq:fromax_admm}) based on ADMM. We relax the linear constraints and form an augmented Lagrangian of (\ref{eq:fromax_admm}) as follows:
\begin{multline}\label{eq:fromax_LA}
L_A(\mathbf{X,Y,q,P}) = \frac{1}{2}\|\mathbf{P - XY}\|_F^2 + \boldsymbol{\boldsymbol{\lambda}}^T (\mathcal{A}(\mathbf{P}) - \mathbf{q}) + \langle \boldsymbol{\Gamma}, \mathbf{Y - X}^T\rangle \\ + \frac{\beta_1}{2} \|\mathcal{A}(\mathbf{P}) - \mathbf{q}\|_2^2 + \frac{\beta_2}{2}\|\mathbf{Y - X}^T\|_F^2+\iota_{\{\mathbf{q}:\|\mathbf{q} - \boldsymbol{1}_S\|_\infty \leq \delta\}}.
\end{multline}
Here, $\boldsymbol{\lambda} \in \mathbb{R}^{N}$ and $\boldsymbol{\Gamma} \in \mathbb{R}^{r \times N}$ represent the scaled Lagrange multipliers. The indicator function, $\iota_{\{\mathbf{q}:\|\mathbf{q} - \boldsymbol{1}_S\|_\infty \leq \delta\}}$, is defined as 
$$\iota_{\{\mathbf{q}:\|\mathbf{q} - \boldsymbol{1}_S\|_\infty \leq \delta\}} = \begin{cases}
0 & \text{ if }\|\mathbf{q} - \boldsymbol{1}_S \|_\infty \leq \delta\\
\infty & \text{ otherwise}.
\end{cases}$$ 
The optimization in (\ref{eq:fromax_LA}) is carried out over $\mathbf{P} \in \mathbb{R}^{N \times N}$, $\mathbf{X} \in \mathbb{R}^{N \times r}$, $\mathbf{Y} \in \mathbb{R}^{r \times N}$, and $\mathbf{q} \in \mathbb{R}^S$, while $\boldsymbol{\lambda}$ and $\boldsymbol{\Gamma}$ are also iteratively updated. We optimize each variable in an alternating fashion like NuMax. The following steps below are performed until convergence.

\textbf{Update $\mathbf{q}$}: Isolating the terms that involve $\mathbf{q}$, we obtain a new estimate $\mathbf{q}_{k + 1}$ as the solution of the constrained optimization problem
\begin{equation}
\mathbf{q}_{k + 1} \leftarrow \underset{\mathbf{q}}{\arg \min } ~ \boldsymbol{\lambda}_k^T (\mathcal{A}(\mathbf{P}_k) - \mathbf{q}) + \frac{\beta_1}{2}\|\mathcal{A}(\mathbf{P}_k) - \mathbf{q} \|_2^2 + \iota_{\{\mathbf{q}:\|\mathbf{q} - \textbf{1}_S\|_\infty \leq \delta\}}.
\end{equation}
Define $\mathbf{z} = \mathcal{A}(\mathbf{P}_k) - \boldsymbol{\lambda}_k - \boldsymbol{1}_S$. Using a component-wise truncation procedure for entries in $\mathbf{q}$, we easily see that
\begin{equation} 
\mathbf{q}_{k + 1} = \boldsymbol{1}_S + \operatorname{sign}(\mathbf{z}) \cdot \min(|\mathbf{z}|, \delta),
\end{equation}
where the sign and min operators are applied component-wise. 

\textbf{Update $\mathbf{P}$}: Isolating the terms that involve $\mathbf{P}$, we obtain a new estimate $\mathbf{P}_{k + 1}$ as the solution of the constrained optimization problem
\begin{equation}
\mathbf{P}_{k + 1} \leftarrow \underset{\mathbf{P}}{\arg \min} ~ \frac{1}{2}\|\mathbf{P} - \mathbf{X}_k \mathbf{Y}_k\|_F^2 + \boldsymbol{\lambda}_k^T (\mathcal{A}(\mathbf{P}) - \mathbf{q}_{k+1}) + \frac{\beta_1}{2} \|\mathcal{A}(\mathbf{P}) - \mathbf{q}_{k + 1}\|_2^2,
\end{equation}
such that $\mathbf{P} \succeq 0$. Since this is a least-squares problem, we can solve for the minimum by solving:
\begin{equation}
(\mathbf{P} - \mathbf{X}_k \mathbf{Y}_k) + \sum_{j = 1}^S \lambda_{k, j} \mathbf{v}_j \mathbf{v}_j^T + \beta_1 \mathcal{A}^*(\mathcal{A}(\mathbf{P}) - \mathbf{q}_{k + 1}) = 0,
\end{equation}
where $\mathcal{A^*}$ is the adjoint of $\mathcal{A}$. The solution is  \begin{equation}
\mathbf{P}_{k+1} = (\mathbf{I} + \beta_1 \mathcal{A} ^*\mathcal{A})^\dag(\beta_1 \mathcal{A}^*\mathbf{q}_{k+1}+\mathbf{X}_k\mathbf{Y}_k-\sum_{j = 1}^s \lambda_{k, j} \mathbf{v}_j \mathbf{v}_j^T)
\end{equation}
where $\dag$ denotes the pseudoinverse. 

\textbf{Update $\mathbf{X}$}: Isolating the terms that involve $\mathbf{X}$, we obtain a new estimate $\mathbf{X}_{k + 1}$ as the solution of the constrained optimization problem
\begin{equation}
\mathbf{X}_{k+1} \leftarrow \underset{\mathbf{X}}{\arg \min} ~ \frac{1}{2}\|\mathbf{P}_{k + 1} - \mathbf{X} \mathbf{Y}_k \|_F^2 + \langle \boldsymbol{\Gamma}_k, \mathbf{Y}_k - \mathbf{X}^T \rangle + \frac{\beta_2}{2}\|\mathbf{Y}_k - \mathbf{X}^T\|_F^2.
\end{equation}

It is easily seen that this can be solved similarly to the $\mathbf{P}$ update.

\textbf{Update $\mathbf{Y}$}: Isolating the terms that involve $\mathbf{Y}$, we obtain a new estimate $\mathbf{Y}_{k + 1}$ as the solution of the constrained optimization problem
\begin{equation}
\mathbf{Y}_{k+1} \leftarrow \underset{\mathbf{Y}}{\arg \min} ~ \frac{1}{2}\|\mathbf{P}_{k + 1} - \mathbf{X}_{k + 1} \mathbf{Y} \|_F^2 + \langle \boldsymbol{\Gamma}_k, \mathbf{Y} - \mathbf{X}_{k + 1}^T \rangle + \frac{\beta_2}{2}\|\mathbf{Y} - \mathbf{X}_{k + 1}^T\|_F^2.
\end{equation}

It is easily seen that this can be solved similarly to the $\mathbf{X}$ update.

\textbf{Update $\boldsymbol{\lambda, \Gamma}$}: Following standard augmented Lagrangian methods, we update $\Gamma, \Pi$ according to the following equations
\begin{align}
\boldsymbol{\lambda}_{k + 1} \leftarrow & \boldsymbol{\lambda}_k + \eta\beta_1(\mathcal{A}(\mathbf{P}_{k+1}) - \mathbf{q}_{k+1})\\
\boldsymbol{\Gamma}_{k+1} \leftarrow  & \boldsymbol{\Gamma}_k + \eta \beta_2(\mathbf{Y}_{k+1} - \mathbf{X}_{k+1}^T).
\end{align}
Pseudocode for FroMax may be found in Algorithm \ref{alg:fromax}. Convergence properties of FroMax are highly dependent on chosen parameters $\eta, \beta_1,$ and $\beta_2$. 

\begin{algorithm}[tb]
   \caption{FroMax}\label{alg:fromax}
\begin{algorithmic}
   \STATE {\bfseries Inputs:} Secant set $\mathcal{S}(\mathcal{X}) = \{\mathbf{v}_i\}_{i = 1}^S$, isometry constant $\delta$, desired rank for $\mathbf{P}$ $r$, max iterations $m > 0$
   \STATE {\bfseries Parameters:} $\beta_1, \beta_2, \eta > 0$
   \STATE {\bfseries Output:} $\mathbf{P}, \mathbf{X}, \mathbf{Y}$\\
   \STATE
   \FOR{$k = 1$ {\bfseries to} $m$}
   \STATE $\mathbf{z} \leftarrow \mathcal{A}(\mathbf{P}_k) - \boldsymbol{\lambda}_k - \boldsymbol{1}_S$
\STATE $\mathbf{q}_{k+1} \leftarrow \boldsymbol{1}_S + \operatorname{sign}(\mathbf{z}) \cdot \min(|\mathbf{z}|,\delta)$
\STATE
\STATE $\mathbf{P}_{k+1} \leftarrow (\mathbf{I} + \beta_1 \mathcal{A} ^*\mathcal{A})^\dag(\beta_1 \mathcal{A}^*\mathbf{q}_{k+1}+\mathbf{X}_k\mathbf{Y}_k-\sum_{j = 1}^s \lambda_{k, j} \mathbf{v}_j \mathbf{v}_j^T)$
\STATE
\STATE $\mathbf{X}_{k+1} \leftarrow (\boldsymbol{\Gamma}_k^T + \beta_2 \mathbf{Y}_k^T + \mathbf{P}_{k+1} \mathbf{Y}_k^T)(\mathbf{Y}_k\mathbf{Y}_k^T + \beta_2 \mathbf{I})^{-1}$
\STATE
\STATE $\mathbf{Y}_{k+1} \leftarrow (\mathbf{X}_{k+1}^T\mathbf{X}_{k+1} + \beta_2 \mathbf{I})^{-1}(\mathbf{X}_{k+1}^T\mathbf{P}_{k+1} - \boldsymbol{\Gamma}_k + \beta_2\mathbf{X}_{k+1}^T)$
\STATE
\STATE $\boldsymbol{\lambda}_{k + 1} \leftarrow \boldsymbol{\lambda}_k + \eta\beta_1(\mathcal{A}(\mathbf{P}_{k+1}) - \mathbf{q}_{k+1})$
\STATE $\boldsymbol{\Gamma}_{k+1} \leftarrow \boldsymbol{\Gamma}_k + \eta \beta_2(\mathbf{Y}_{k+1} - \mathbf{X}_{k+1}^T)$
\STATE
\IF{$\frac{1}{2}\|\mathbf{P}_{k+1} - \mathbf{X}_{k+1}\mathbf{Y}_{k+1}\|_F^2 < \epsilon$}
\STATE break
\ENDIF
\STATE
   \ENDFOR
   \RETURN $\mathbf{P} \leftarrow \mathbf{P}_k, \mathbf{X} \leftarrow \mathbf{X}_k, \mathbf{Y} \leftarrow \mathbf{Y}_k$
\end{algorithmic}
\end{algorithm}

\section{NILE-Pro} \label{nile-pro}

Our second algorithm, \textit{Near-Isometric Linear Embedding via Proximal Mapping}, or \textit{NILE-Pro} seeks to minimize the RIP constraint directly to solve for $\boldsymbol{\Psi}$. This minimization problem is solved using ADMM and a Moreau decomposition on a proximal mapping.

\subsection{Optimization Framework}

We formulate a new framework for NILE-Pro. We solve for our desired linear embedding $\boldsymbol{\Psi}$ directly:
\begin{equation}\label{eq:nile-pro}
\underset{\boldsymbol{\Psi}}{\text{minimize }} \|\mathcal{A}(\boldsymbol{\Psi}^T\boldsymbol{\Psi})-\boldsymbol{1}_S\|_\infty.
\end{equation}

By introducing an auxiliary variable $\mathbf{q}$, which simplifies our problem and allows us to apply constrained optimization techniques, we then have the following minimization problem:
\begin{equation}
\begin{aligned}
\underset{\mathbf{q}, \boldsymbol{\Psi}}{\text{minimize }} & \|\mathbf{q} - \boldsymbol{1}_S\| _\infty\\
\text{subject to }&  \mathbf{q} =  \mathcal{A}(\boldsymbol{\Psi}^T\boldsymbol{\Psi}).
\end{aligned}
\end{equation}
We apply ADMM and use a Moreau decomposition on a proximal mapping to solve for updates. Like FroMax, this optimization problem is non-convex and thus, we cannot guarantee that NILE-Pro will converge to the optimal solution of (\ref{eq:nile-pro}). However, we demonstrate in \S \ref{experiments} that NILE-Pro may produce stable, excellent results for synthetic and real-world data sets at a much faster rate than both FroMax and NuMax due to the simplified problem it solves.

\subsection{Proximal Mapping and Moreau Decomposition}

We introduce some machinery to solve this minimization problem:

\begin{definition} 
The \textit{proximal mapping} \cite{Rockafellar70convexanalysis} of a closed proper convex function $f$ is defined to be $$\prox_f(\mathbf{x}) = \underset{\mathbf{u}}{\arg\min} ( f(\mathbf{u}) + \frac{1}{2}\|\mathbf{u - x}\|_2^2).$$
\end{definition}

The proximal mapping may be interpreted as a compromise between minimizing $f$ and staying near the original iterate $x$. Proximal mappings generalize projections, and are particularly useful since they often work very fast in practice.\\

\begin{theorem} 
If a function $f:\mathbb{R}^n \rightarrow \mathbb{R}$ is proper, closed, and convex, then $\prox_f(\mathbf{x})$ exists, well-defined, and unique for all $\mathbf{x}$ \cite{Rockafellar70convexanalysis}.\\
\end{theorem}

This theorem guarantees that these proximal mappings exist for closed, proper, and convex functions.\\

\begin{definition} 
The \textit{convex conjugate} \cite{Rockafellar70convexanalysis} of a closed proper convex function $f$ is defined to be $$f^*(\mathbf{y}) = \underset{\mathbf{x}}{\sup} (\mathbf{y}^T \mathbf{x} - f(\mathbf{x})).$$
\end{definition}

Then for any $\mathbf{x}$, the following relation holds: $$\mathbf{x} = \prox_f(\mathbf{x}) + \prox_{f^*}(\mathbf{x}),$$ where $f^*$ is the convex conjugate of $f$. This decomposition, called the \textit{Moreau decomposition}, generalizes the orthogonal decomposition on subspaces. We apply this machinery to help us solve for the update for $\mathbf{q}$.

\subsection{ADMM}

Following a similar method as FroMax, we relax our linear constraints and find our augmented Lagrangian of (14): 
\begin{equation}\label{eq:nile-pro_LA}
L_A(\boldsymbol{\Psi}, \mathbf{q}; \boldsymbol{\omega}) = \|\mathbf{q} - \boldsymbol{1}_S\|_\infty + \frac{\beta}{2} \|\mathcal{A}(\boldsymbol{\Psi}^T\boldsymbol{\Psi}) - \mathbf{q} - \boldsymbol{\omega} \|_2^2.
\end{equation}
Here, $\boldsymbol{\omega} \in \mathbb{R}^{S}$ is the scaled Lagrange multiplier. The optimization in (\ref{eq:nile-pro_LA}) is carried out over $\boldsymbol{\Psi} \in \mathbb{R}^{r \times N}$ and $\mathbf{q} \in \mathbb{R}^S$, while $\boldsymbol{\omega}$ is updated. Each variable is updated in an alternating fashion. The following steps below are performed until convergence.

\textbf{Update $\boldsymbol{\Psi}$}: Isolating the terms that involve $\boldsymbol{\Psi}$, we obtain a new estimate $\boldsymbol{\Psi}_{k + 1}$ as the solution of the constrained optimization problem
\begin{equation}
\boldsymbol{\Psi}_{k + 1} \leftarrow \underset{\boldsymbol{\Psi}}{\arg\min} ~ \frac{\beta}{2} \| \mathcal{A}(\boldsymbol{\Psi}^T\boldsymbol{\Psi}) - \mathbf{q}_{k} - \boldsymbol{\omega}_k \|_2^2.
\end{equation}

\textbf{Update $\mathbf{q}$}: Isolating the terms that involve $\mathbf{q}$, we obtain a new estimate $\mathbf{q}_{k + 1}$ as the solution of the constrained optimization problem
\begin{equation}
\mathbf{q}_{k + 1} \leftarrow \underset{\mathbf{q}}{\arg\min} ~ \|\mathbf{q} - \boldsymbol{1}_S \|_\infty +  \frac{\beta}{2} \| \mathcal{A}(\boldsymbol{\Psi}_{k + 1}^T\boldsymbol{\Psi}_{k + 1}) - \mathbf{q} - \boldsymbol{\omega}_k \|_2^2.
\end{equation}
Set $\mathbf{x} = \mathbf{q}_{k + 1} - \boldsymbol{1}_S$ and $\boldsymbol{\tau} = \mathcal{A}(\boldsymbol{\Psi}_{k + 1}^T\boldsymbol{\Psi}_{k + 1}) - \boldsymbol{\omega}_k - \boldsymbol{1}_S$, let $f(\mathbf{x}) = \frac{1}{\beta}\|\mathbf{x}\|_\infty$ and its convex conjugate $$f^*(\mathbf{x})=\left\{
		\begin{array}{ll}
			0 & \mbox{if } \|\mathbf{\beta x}\|_1 \leq 1 \\
			\infty & \mbox{if } \|\mathbf{\beta x}\|_1 > 1
		\end{array}
	\right\}.$$ Then $\prox_{f^*}(\boldsymbol{\tau})= \mathcal{P}_{\{\|x\|_1 \leq 1\}}(\beta \boldsymbol{\tau})$, i.e. the projection of $\beta\boldsymbol{\tau}$ to the $\ell^1$ ball. Then applying Moreau's decomposition, we have the $\mathbf{q}$ update:
\begin{align}
\mathbf{x} & = \frac{1}{\beta}(\beta \boldsymbol{\tau} - \mathcal{P}_{\{\|x\|_1 \leq 1\}}(\beta \boldsymbol{\tau}))\\
\mathbf{q} & = \frac{1}{\beta}(\beta \boldsymbol{\tau} - \mathcal{P}_{\{\|x\|_1 \leq 1\}}(\beta \boldsymbol{\tau})) + \boldsymbol{1}_S.
\end{align}

\textbf{Update $\boldsymbol{\omega}$}: Following standard augmented Lagrangian methods, we update $\boldsymbol{\omega}$ according to the following equation
\begin{equation}
\boldsymbol{\omega}_{k + 1} \leftarrow \boldsymbol{\omega}_k - \beta(\mathcal{A}(\boldsymbol{\Psi}^T_{k + 1}\boldsymbol{\Psi}_{k + 1}) - \mathbf{q}_{k + 1}).
\end{equation}

Pseudocode for NILE-Pro may be found in Algorithm \ref{alg:nile-pro}.

\begin{algorithm}[tb]
\caption{NILE-Pro}\label{alg:nile-pro}
\begin{algorithmic}
\STATE {\bfseries Inputs:} Secant set $\mathcal{S}(\mathcal{X}) = \{\mathbf{v}_i\}_{i = 1}^S$, isometry constant $\delta$, max iterations $m > 0$, initial rank $r$
\STATE {\bfseries Parameters:} $\beta > 0$
\STATE {\bfseries Output:} $\boldsymbol{\Psi}$
\STATE
\FOR{$k = 1, ..., m$}
\STATE $\tau \leftarrow \mathcal{A}(\boldsymbol{\Psi}_k^T\boldsymbol{\Psi}_k) -\boldsymbol{\omega} - 1_s $
\STATE $\mathbf{q}_{k+1} \leftarrow\frac{1}{\beta} \Big(\beta\tau-\mathcal{P}_{\{\|x\|_1 \leq 1\}}(\beta\tau) \Big) +\boldsymbol{1}_S$
\STATE
\STATE $\boldsymbol{\Psi}_{k+1} \leftarrow \boldsymbol{\Psi}_k - 2\eta \boldsymbol{\Psi}_k \mathcal{A}^*(\mathcal{A}(\boldsymbol{\Psi}_k^T\boldsymbol{\Psi}_k) - \mathbf{q}_{k+1}-\boldsymbol{\omega})$
\STATE
\STATE $\boldsymbol{\omega}_{k + 1} \leftarrow \boldsymbol{\omega}_k - \beta(\mathcal{A}(\boldsymbol{\Psi}_{k+1}^T\boldsymbol{\Psi}_{k+1}) - \mathbf{q}_{k+1})$
\STATE
\STATE $\epsilon_0 \leftarrow \|\mathcal{A}(\boldsymbol{\Psi}_{k+1}^T\boldsymbol{\Psi}_{k+1})- \boldsymbol{1}_S\|_\infty$\\
\IF{$ \epsilon_0 < \epsilon$}
\STATE break
\ENDIF
\STATE
\ENDFOR
\RETURN $\boldsymbol{\Psi} \leftarrow \boldsymbol{\Psi}_k$
\end{algorithmic}
\end{algorithm}

\section{Rank Adjustment and Column Generation} \label{racg}

In this section, we discuss rank adjustment and column generation heuristics. We develop rank adjustment methods to discover the lowest optimal rank for both FroMax and NILE-Pro. Abbreviating rank adjustment to \textit{RA}, we call our rank adjusted algorithms \textit{FroMax RA} and \textit{NILE-Pro RA}, respectively. We also use column generation techniques following Hegde et. al. \cite{HegdeSankaranarayananYinBaraniuk2015_numax} to work with subsets of $S(\mathcal{X})$ to lower the memory complexity of these algorithms, which we name \textit{FroMax CG} and \textit{NILE-Pro CG}, respectively. We discuss each heuristic algorithm in detail below.

\subsection{Rank Adjustment}

Though FroMax and NILE-Pro may dramatically decrease the time of solving for projection matrix $\boldsymbol{\Psi}$, both algorithms do not find an optimal rank for dimensionality reduction like NuMax. Hence, we propose a heuristic rank adjustment method that uses the discovered matrix $P = \boldsymbol{\Psi}^T\boldsymbol{\Psi}$ to give a good initialization for $\boldsymbol{\Psi}$ of lower rank.

Given a sufficiently large rank, $R_0 \gg r$, the optimal rank, we run our dimensionality reduction algorithm for a maximum number of iterations or until convergence to find $\boldsymbol{\Psi}$. If our algorithm converges, we return $\mathbf{P} = \boldsymbol{\Psi}^T\boldsymbol{\Psi}$ and find $\boldsymbol{\Psi}_0 = \boldsymbol{\Gamma}_M^{1/2}\mathbf{U}_M^T$, where $\mathbf{P} = \mathbf{U} \boldsymbol{\Gamma} \mathbf{U}^T$ from $\mathbf{P}$'s eigendecomposition. We then initialize our algorithm again with rank $R_1 = R_0-1$ and $\boldsymbol{\Psi}_0$ which we found in the last iteration and test again for convergence.  We continue this process until we reach the maximum number of iterations within the algorithm and return the $\boldsymbol{\Psi}$ given in the last iteration, considering its rank $r = R_k$ to be optimal. We summarize our rank adjustment heuristic in Algorithm \ref{alg:RA}.

\begin{algorithm}[tb]
\caption{FroMax/NILE-Pro RA}\label{alg:RA}
\begin{algorithmic}
\STATE {\bfseries Inputs:} Secant set $\mathcal{S}(\mathcal{X}) = \{\mathbf{v}_i\}_{i = 1}^S$, isometry constant $\delta$, max iterations for algorithm $m > 0$, initial rank $R_0$, max iterations for RA $M$, $\boldsymbol{\Psi}_0$
\STATE
\FOR{k = 1, ..., M}
\STATE $\boldsymbol{\Psi} \leftarrow \operatorname{FroMax/NILE-Pro}(S, \delta, m, R_0, \boldsymbol{\Psi}_0)$
\STATE $P \leftarrow \boldsymbol{\Psi}^T\boldsymbol{\Psi}$
\STATE $R_{k + 1} \leftarrow R_k - 1$
\STATE $[\Gamma, U] \leftarrow \operatorname{eig}(P)$
\STATE $\boldsymbol{\Psi}_0 \leftarrow \Gamma^{1/2}U^T$
\STATE
\IF{FroMax/NILE-Pro does not converge}
\STATE break
\ENDIF
\STATE
\ENDFOR
\STATE return $\boldsymbol{\Psi}, r \leftarrow R_k$
\end{algorithmic}
\end{algorithm}

\subsection{Column Generation}

Since FroMax and NILE-Pro use the secants of a given data set, applications involving millions of secants may be prohibited by the memory complexity of these algorithms. Some methods that are used to address large data sets include stochastic and online methods. Stochastic methods use random subsets of the data to learn an estimate for the entire data set \cite{kushner2003stochastic, spall2005introduction}. Online methods uses sequentially available data to update the current iterate then discards the information \cite{Mairal:2010:OLM:1756006.1756008,le2004large}. Our column generation algorithms, FroMax CG and NILE-Pro CG, combines stochastic and online methods to estimate solutions to large-scale problems.

Similar to NuMax's column generation, which is based off of the Karush-Kuhn-Tucker (KKT) conditions, we apply a simple, greedy method to rapidly find the active constraints for (\ref{eq:fromax}) or (\ref{eq:nile-pro}).
\begin{enumerate}
\item Solve (\ref{eq:fromax}) or (\ref{eq:nile-pro}) with a small subset $S_0 \subset S(\mathcal{X})$ using FroMax (Algorithm \ref{alg:fromax}) or NILE-Pro (Algorithm \ref{alg:nile-pro}), respectively to obtain an initial estimate $\widehat{\boldsymbol{\Psi}}$. Identify the set $\widehat{S}$ of secants that correspond to the active constraints: $$\widehat{S} \leftarrow \{ \mathbf{v}_i \in S_0 : |\mathbf{v}_i^T\widehat{\boldsymbol{\Psi}}^T \widehat{\boldsymbol{\Psi}} \mathbf{v}_i - 1| \geq \delta \}.$$
\item Select additional secants $S_1 \subset S(\mathcal{X})$ not selected previously and identify all secants among $S_1$ that violate the constraint at the current estimate $\widehat{\boldsymbol{\Psi}}$. Then, append these secants to the set of active constraints $\widehat{S}$ to obtain an augmented set $\widehat{S}$ $$\widehat{S} \leftarrow \widehat{S} \bigcup \{ \mathbf{v}_i \in S_1 : |\mathbf{v}_i^T\widehat{\boldsymbol{\Psi}}^T \widehat{\boldsymbol{\Psi}}\mathbf{v}_i - 1 | \geq \delta \}.$$
\item Solve (\ref{eq:fromax}) or (\ref{eq:nile-pro}) with the new augmented set $\widehat{S}$ using FroMax or NILE-Pro to obtain a new estimate $\widehat{\boldsymbol{\Psi}}$.
\item Identify the secants that correspond to active constraints and repeat Steps 2 and 3 until convergence is reached for $\widehat{\boldsymbol{\Psi}}$.
\end{enumerate}

Column generation allows us to perform a large numerical optimization procedure on smaller subsets of $S(\mathcal{X})$, resulting in significant computational gains. A key benefit of FroMax CG and NILE-Pro CG is that the subsets of secants used during each iteration never has to be explicitly stored in memory and can be generated on the fly. This leads to significant improvements in memory complexity. 

However, because FroMax and NILE-Pro are already both non-convex, column generation makes these algorithms even less predictable. Though these algorithms are not guaranteed to converge to an optimal solution, they appear to yield excellent results on large, real-world data sets, as we will show in \S \ref{experiments}. 

Pseudocode for FroMax/Nile-Pro CG is found in Algorithm \ref{alg:CG}. Our column generation method converges when no additional secants violate our constraint.

\begin{algorithm}[tb]
\caption{FroMax/NILE-Pro CG}\label{alg:CG}
\begin{algorithmic}
\STATE {\bfseries Inputs:} Secant set $\mathcal{S}(\mathcal{X}) = \{\mathbf{v}_i\}_{i = 1}^S$, isometry constant $\delta$, max iterations for algorithm $m > 0$, rank $r$, the FroMax or NILE-Pro algorithm
\STATE
\WHILE{not converged}
\STATE $\widehat{S} \leftarrow \{\mathbf{v}_i \in S_0 : |\mathbf{v}_i^T\boldsymbol{\Psi}^T\boldsymbol{\Psi} \mathbf{v}_i - 1| \geq \delta \}$
\STATE $S_1 \leftarrow \{\mathbf{v}_i \in S(\mathcal{X}) : \mathbf{v}_i \notin S_0 \}_{i = 1}^{S''}$
\STATE $\widehat{S} \leftarrow \widehat{S} \bigcup \{\mathbf{v}_i \in S_1 : |\mathbf{v}_i^T \boldsymbol{\Psi}^T \boldsymbol{\Psi} \mathbf{v}_i = 1| \geq \delta \}$\\
\STATE $\boldsymbol{\Psi} \leftarrow \operatorname{FroMax/NILE-Pro}(\widehat{S}, \delta)$
\STATE $S_0 \leftarrow \widehat{S}$
\ENDWHILE
\STATE
\STATE return $\boldsymbol{\Psi}$
\end{algorithmic}
\end{algorithm}

\section{Convergence of Algorithms}

Since FroMax and NILE-Pro are derived from applying the ADMM to non-convex problems, the convergence properties of these algorithms can be understood based on the known convergence properties of ADMM. For certain types of convex problems, ADMM has been shown to converge at a rate of $o(1/k)$ \cite{deng20131}. Recent results by Hong, et.al. and Wang, et. al. \cite{hong2015convergence,wang2015global} show that ADMM does converge for certain classes or types of non-convex problems. These results may suggest convergence of our algorithms.

\section{Numerical Experiments} \label{experiments}

We demonstrate the performance of the FroMax and NILE-Pro algorithms in comparison to prior methods including NuMax. All of our experiments are performed on computers with Intel i5-650 processors and 4 GB of RAM unless otherwise specified. We test and compare the speed and accuracy of our algorithms through various tests on real-world and synthetic data sets.

\subsection{Linear Low-Dimensional Embeddings}\label{linear1}

We first consider synthetic data sets $\mathcal{X}_1$ and $\mathcal{X}_2$ consisting of $7 \times 7 = 49$ and $14 \times 14 = 196$ dimensional images of translations of a white square on a black box respectively. We construct our training sets by randomly generating $60$ $49$-dimensional images for $\mathcal{X}_1$ and $200$ $196$-dimensional images for $\mathcal{X}_2$. We then construct secant sets $S(\mathcal{X}_1)$ and $S(\mathcal{X}_2)$ by computing the normalized pairwise difference vectors between different images. We compare FroMax and NILE-Pro's performance of producing linear, low-dimensional embeddings on these two data sets in Table 7.1.

\begin{table}[t]
\begin{center}
\begin{small}
\begin{sc}
\begin{tabular}{lc c c c c r}
\hline
& & \multicolumn{2}{c}{FroMax} & \multicolumn{2}{c}{NILE-Pro}\\
\#Data & $\delta$ & Rank & Time & Rank & Time\\
\hline  
\multirow{3}{*}{60} & 0.4  & 9 & 1.3 & 9 & 0.7\\
& 0.25 & 9 & 1.2 & 9 & 1.1\\
& 0.1 & 13 & 1.4 & 13 & 1.5\\
\hline  
\multirow{3}{*}{200} & 0.4  & 16 & 511.2 & 16 & 109.1\\
& 0.25 & 18 & 269.4 & 18 & 144.4\\
& 0.1 & 27 & 74.5 & 27 & 448.5\\
\hline
\end{tabular}
\end{sc}
\end{small}
\end{center}
\caption{Comparison of runtime performance for FroMax and NILE-Pro on $S(\mathcal{X}_1)$ and $S(\mathcal{X}_2)$ given $\delta$ and the optimal rank from Numax RA.}
\end{table}

Since NILE-Pro minimizes the RIP directly, NILE-Pro intuitively will converge faster for larger $\delta$. Our experimental results match our theoretical intuition since NILE-Pro converges significantly faster for larger $\delta$ than lower $\delta$. 

FroMax experimentally converges faster for smaller $\delta$ than larger $\delta$. Smaller $\delta$ restricts $q$ to a smaller feasible set given by the RIP, leading to faster convergence.

Also note that both algorithms' computational complexity scale significantly with the size of the data set due to the use of the secant set. Our runtime results comparing $S(\mathcal{X}_1)$ and $S(\mathcal{X}_2)$ reflect this.

\subsection{Linear Low-Dimensional Embeddings with Rank Adjustment}

\begin{figure}[ht!]
\centering
\includegraphics[width=2in]{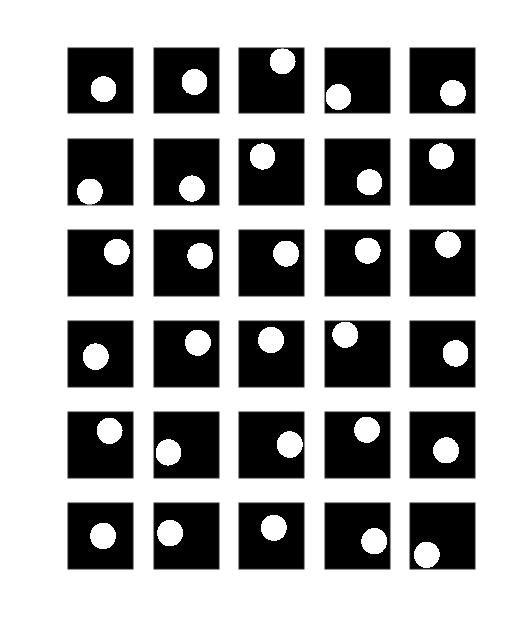}
\includegraphics[width=2in]{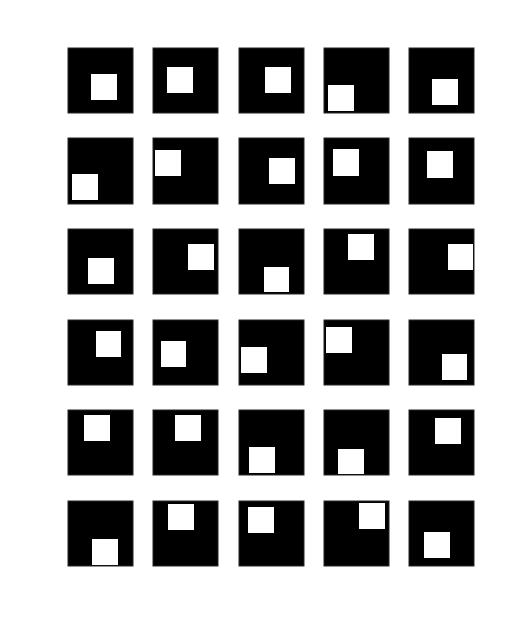}
\caption[Sample Disk Figure]{Our synthetic training set consists of sixty $256$-dimensional random generated translating disks and squares figure.}
\label{fig:disk}
\end{figure}

In \S \ref{linear1}, we input a given rank for FroMax and NILE-Pro and compare their run time. However, usually the optimal rank for dimension reduction is not known, motivating the development of rank adjustment heuristics. To analyze the performance of our rank adjustment heuristic, we consider a synthetic data set $\mathcal{X}$ comprised of $16 \times 16 = 256$ dimensional images of translations of a white square or a disk on a black box respectively, see figure~\ref{fig:disk}. We construct a secant set $S(\mathcal{X})$ and compare PCA, Numax RA, FroMax RA and NILE-Pro RA's performance of producing linear, low-dimensional embeddings on this data set.

\begin{figure}[ht!]

\hspace{0.5in}
\includegraphics[scale = .36]{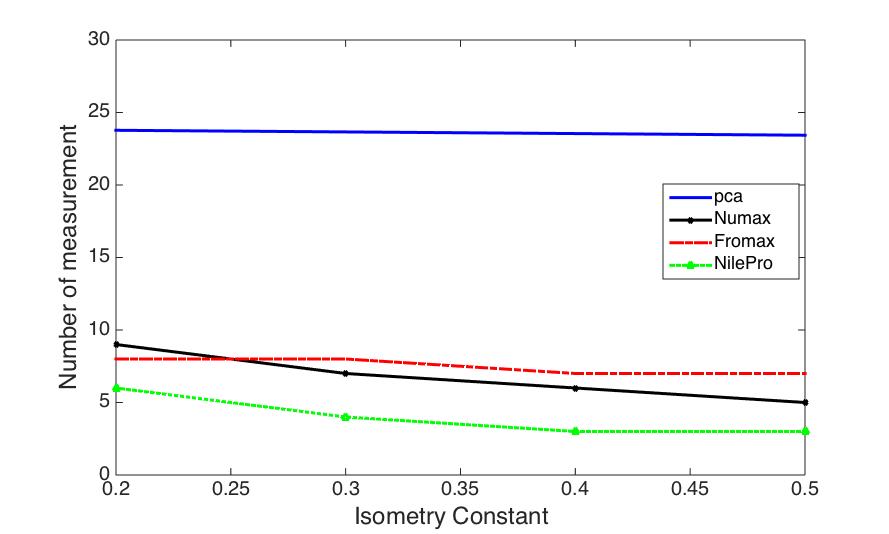}
\caption{A comparison of the isometry constant $\delta$ with the number of measurements for PCA, Numax RA, FroMax RA and NILE-Pro RA's performance of producing linear, low-dimensional embeddings. }
\label{fig:bw}
\end{figure}

Figure~\ref{fig:bw} plots the variation of the number of measurements $M$ as a function of the isometry constant $\delta$. We observe that NILE-Pro RA achieves the desired isometry constant on the secants using by far the fewest number of measurements. FroMax RA performs better for small $\delta$ due to the correlation between $\delta$ and $q$, as we discussed before. Moreover, both Numax RA and Fromax RA greatly outperform PCA, a popular embedding technique in the literature. 

\subsection{Runtime Performance on MNIST with Rank Adjustment}

\begin{figure}[ht!]
\centering
\includegraphics[width=2in]{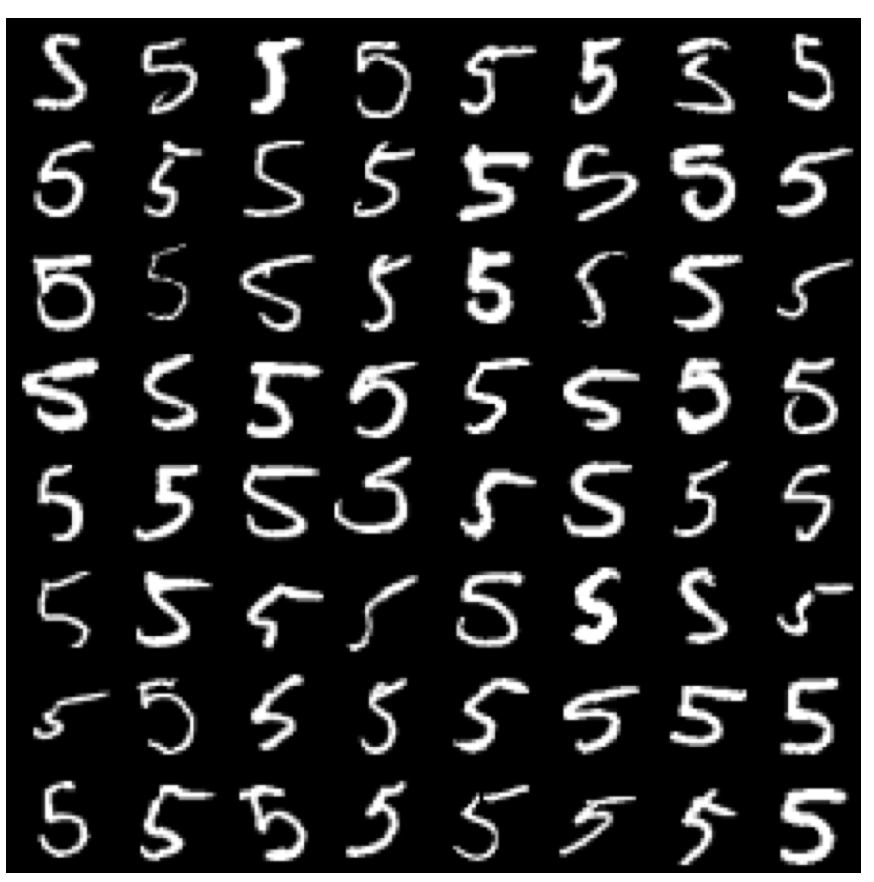}
\caption{Examples of “5” images from the MNIST dataset.}
\label{fig:MNIST}
\end{figure}

In this experiment, we consider a more challenging, real-world data set, the MNIST data set, see figure~\ref{fig:MNIST}. MNIST contains many digital images of handwritten digits and is a common benchmark data set for machine learning. We examine subsets of the training set for the digit ``5''. We take subsets consisting of 95, 200, and 500 data points with original dimension 49.

We test runtime performance of FroMax and NILE-Pro RA on these data sets. Our results may be found in Table 2.

\begin{table}[t]
\begin{center}
\begin{small}
\begin{sc}
\begin{tabular}{l c c c c c c c r}
\hline  
& & \multicolumn{2}{c}{NILE-Pro} & \multicolumn{2}{c}{FroMax} & \multicolumn{2}{c}{NuMax}\\
$\delta$ & \#Data & Rk & Time & Rk & Time & Rk & Time\\
\hline  
\multirow{3}{*}{0.4} & 95 & 7 & 25 & 9 & 102 & 12 & 71\\
 & 200 & 9 & 96 & 15 & 520 & 21 & 311\\
 & 500 & 11 & 710 & 27 & 2490 & 25 & 3477\\
\hline  
\multirow{3}{*}{0.2} & 95 & 11 & 28 & 11 & 111 & 14 & 56\\
 & 200 & 14 & 130 & 16 & 569 & 18 & 557\\
 & 500 & 18 & 751 & 40 & 1498 & 27 & 3517\\
\hline  
\multirow{3}{*}{0.1} & 95 & 15 & 41 & 15 & 91 & 16 & 21\\
 & 200 & 20 & 165 & 19 & 823 & 21 & 279\\
 & 500 & 25 & 1285 & 44 & 650 & 30 & 3410\\
\hline
\end{tabular}
\end{sc}
\end{small}
\end{center}
\caption{Comparison of runtime performance for FroMax RA, NILE-Pro RA, and NuMax on subsets of ``5'' images from MNIST.}
\end{table}

Our experimental results show that NILE-Pro RA may perform significantly faster and give a better optimal rank than NuMax while FroMax RA converges slower for larger data sets. This may be due to the nature of the local minima found in FroMax; the estimate for $P = \boldsymbol{\Psi}^T \boldsymbol{\Psi}$ given for a larger rank does not correspond to the local minima for lower ranks so that this initialization is beneficial.

Our results for FroMax RA reveal another issue with our rank adjustment method. FroMax RA appears to struggle with determining the optimal rank, sometimes performing worse than NuMax. We believe that our algorithm may be converging to local minima, which makes our rank adjustment ineffective. This issue motivates us to look into other rank adjustment methods that start at a sufficiently low rank and examine higher ranks to discover the optimal rank. 

Also, since rank adjustment for NILE-Pro is still based on the core NILE-Pro algorithm, we see that NILE-Pro RA converges in much slower time for smaller $\delta$.

The former caveat motivates us to continue looking for better rank adjustment methods for FroMax.

%
%


\subsection{Nearest Neighbor Classification on MNIST}

The MNIST data set consists of 60,000 training data points and 10,000 test data points of handwritten digits \cite{lecun1998mnist}. The dataset contains 10 classes corresponding to each digit from 0-9. For this experiment, we use the $N = 20 \times 20 = 400$-dimensional data set that excludes extra space at the boundaries. We use NuMax CG and FroMax CG to embed our MNIST training set into a lower dimensional space and nearest neighbor classification.

The misclassification rate of nearest neighbor classification on the unchanged data set is 3.47\%. Table \ref{table:classification} gives the nearest neighbor classification misclassification rate for NuMax CG and FroMax CG for given $\delta$ and rank applied on the MNIST data set. In particular, though NuMax CG and FroMax CG give similar misclassification rates, FroMax CG has significantly better runtime performance than NuMax CG. Though a combined rank adjustment and column generation method has not been implemented for FroMax, the results suggest that FroMax may find a sufficiently good projection matrix in much less time.

\begin{table}[t]
\begin{center}
\begin{small}
\begin{sc}
\begin{tabular}{l c c c c c}
\hline
$\delta$ & Rank & NuMax CG & Time (hrs) & FroMax CG & Time (hrs)\\
\hline
0.4 & 72 & 3.09\% & 0.926 & 3.00\% & 0.411\\
0.25 & 98 & 3.15\% & 1.273 & 3.26\% & 0.811\\
0.1 & 167 & 3.31\% & 2.664 & 3.42\% & 1.358\\
\hline
\end{tabular}
\end{sc}
\end{small}
\end{center}
\caption{Comparison of misclassification rates and run-time performance of approximate nearest neighbor classifiers using NuMax CG and FroMax CG for given $\delta$ and rank on the MNIST test set.} \label{table:classification}
\end{table}

\subsection{Approximate Nearest Neighbors}

Given a data set modeled by points in Euclidean space and a query point, \textit{nearest neighbors} identifies the $k$ closest points in the data set \cite{cover1967nearest}. These points are usually used for further processing, such as unsupervised or supervised regression and classification.

However, as the dimension $N$ of the data set grows, the computational cost of identifying the $k$ nearest neighbors also becomes increasingly expensive. An alternative to computing nearest neighbors directly is to embed the data into a lower-dimensional subspace while preserving near-isometry, then applying nearest neighbor techniques. This method is called \textit{approximate nearest neighbors}. Since NuMax, FroMax, and NILE-Pro construct low rank, near-isometric linear embeddings for a given distortion $\delta$, they may potentially enable efficient ANN computations for high-dimensional data sets.

For this experiment, we use the LabelMe data set consisting of 4000 images of indoor and outdoor scenes \cite{russell2008labelme}. We then computed GIST descriptors for each image, which are vectors of size $N = 512$ that roughly describe the overall spatial statistics of the image \cite{oliva2001modeling}. We then used NuMax CG and FroMax CG to estimate low rank, near-isometric linear embeddings for this data set for a given distortion parameter $\delta$. Then we perform ANN computations on 1000 test data points in the corresponding low dimensional space. We compute embeddings of various ranks for FroMax CG to compare performance between different ranks.

\begin{figure}
\includegraphics[scale=0.45]{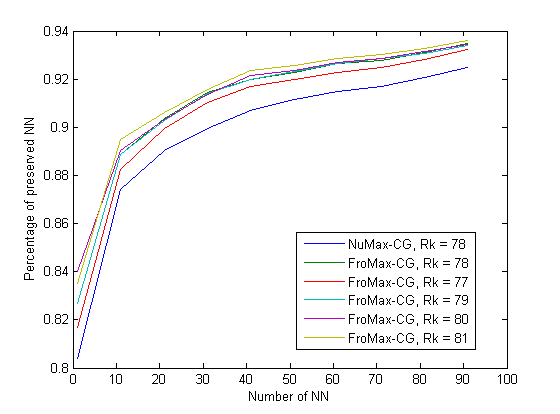}
\includegraphics[scale=0.45]{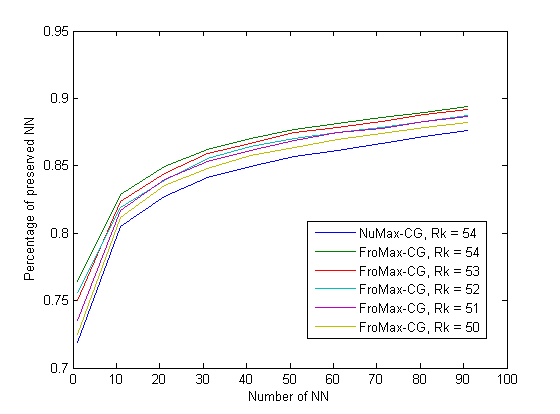}\\
\includegraphics[scale=0.45]{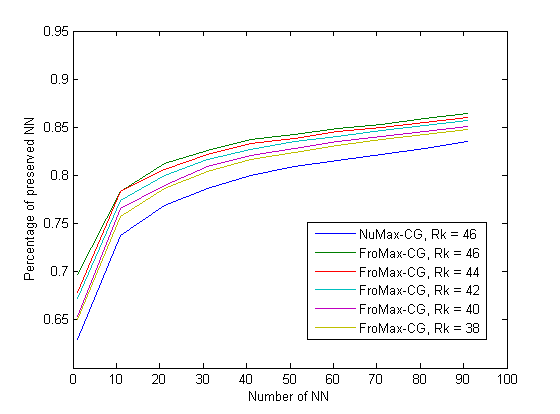}
\includegraphics[scale=0.45]{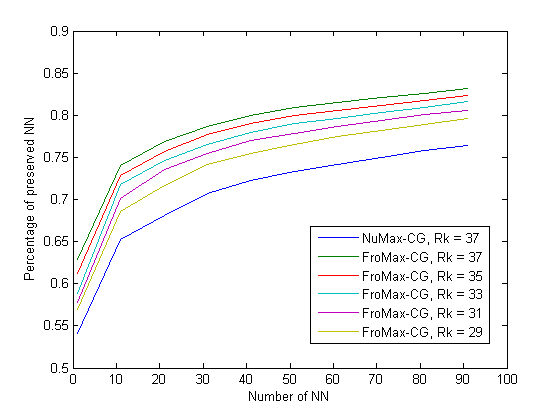}
\caption{Comparison of FroMax CG and NuMax CG on preserving nearest neighbors.}
\label{fig:ann}
\end{figure}

Figure \ref{fig:ann} demonstrates that FroMax CG generally attains similar if not better performance than NuMax CG for the same rank. In fact, our results suggest that FroMax CG could perform similarly at a lower rank than NuMax CG. We leave further investigation for future research.

\section{Discussion} \label{discussion}

\subsection{Research Overview}

In this paper, we construct two comprehensive algorithmic frameworks for finding near-isometric linear embeddings of high-dimensional data sets. Based on the convex optimization formulation in NuMax, we proposed two non-convex minimization approaches which approximately preserve the norms of all pairwise secants of the given dataset. In particular, we developed two algorithms, FroMax and NILE-Pro, that may construct the desired embedding with smaller computational complexity than NuMax. 

Since NuMax automatically discovers the optimal rank, we created a rank adjustment method for finding the best rank for our algorithms. We also implemented column generation in addition to FroMax and NILE-Pro so our algorithms can be adapted to perform on larger data sets. 

Constructing linear, information-preserving embeddings of high-dimensional signals to lower-dimensional signals have become of significant importance for a wide range of machine learning and compressive sensing applications. However, little is known about near-isometric linear embeddings beyond the Johnson-Lindenstrauss Lemma. The frameworks discussed in this paper build on the convex, deterministic approach of NuMax to produce practical, potentially more computationally efficient dimension reduction algorithms that are both information-preserving and feasible for a broad range of applications. Though we do not provide an analytical foundation to our work due to the non-convex nature of our algorithms, we hope to initiate work in developing a theoretical basis for similar work.

\subsection{Future Work}

There are still many challenges left to tackle. As discussed in \S \ref{experiments}, we still need to further develop rank adjustment methods for FroMax and column generation techniques for NILE-Pro. We would also like to incorporate both rank adjustment and column generation together. One direction is to consider an eigengap heuristic for rank adjustment, in which some heuristic is set based on the difference between singular values of the matrix $P$ to determine the next chosen rank.

In addition, further testing and parameter-tweaking is necessary to analyze and optimize the stability and performance of our algorithms on various data sets. Other heuristics for non-convex optimization, such as applying perturbations to avoid local minima, may also be applied to give better solutions. We defer the study of these challenges and heuristics for future research.

\section*{Acknowledgements}

The work of Jerry Luo, Kayla Shapiro, and Hao-Jun Michael Shi were supported in part by the California Research Training Program for Computational and Applied Mathematics 2014 under NSF Grant DMS-1045536. The work of Qi Yang was supported in part by USC Provost's Undergraduate Research Fellowship and the WiSE Research Experience for Undergradutes. We would like to thank Dr. Ming Yan and Dr. Wotao Yin for their consistent support and mentorship throughout the entirety of this project.

\bibliographystyle{plain}
\bibliography{bibl}

\end{document}